\documentclass[letterpaper, 10 pt, conference]{ieeeconf}  

\IEEEoverridecommandlockouts                              

\usepackage{makecell}
\usepackage{graphicx}
\usepackage{xcolor}
\usepackage{soul}
\usepackage[ruled,vlined]{algorithm2e}
\usepackage{amsmath}
\usepackage{gensymb}
\usepackage{bbm}
\usepackage{multirow}
\usepackage{hyperref}
\usepackage{booktabs}
\usepackage{subfigure}
\title{\LARGE \bf
Context-aware Sparse Spatiotemporal Learning for Event-based Vision}

\author{Shenqi Wang and Guangzhi Tang
\thanks{Shenqi Wang is with the Faculty of Aerospace Engineering, Delft
University of Technology, Delft 2628 CD, The Netherlands.
        {\tt\small s.wang-18@tudelft.nl} $^\dag$Corresponding author}%
\thanks{Guangzhi Tang is with the Department of Advanced Computing Sciences, Maastricht University, Maastricht 6211 LK, The Netherlands.
{\tt\small guangzhi.tang@maastrichtuniversity.nl}}%
\thanks{This publication is part of the project Brain-inspired MatMul-free Deep Learning for Sustainable AI on Neuromorphic Processor with file number NGF.1609.243.044 of the research programme AiNed XS Europe which is (partly) financed by the Dutch Research Council (NWO) under the grant https://doi.org/10.61686/MYMVX53467.}%
\thanks{Code of the paper is available in \url{https://github.com/ERNIS-LAB/CSSL-Event-Object-Detection}}
}

\usepackage{tikz}
\usepackage{textcomp}
\usepackage{hyperref}
\usepackage{lipsum}

\newcommand\copyrighttext{%
  \footnotesize \textcopyright 2025 IEEE. Personal use of this material is permitted.
  Permission from IEEE must be obtained for all other uses, in any current or future
  media, including reprinting/republishing this material for advertising or promotional
  purposes, creating new collective works, for resale or redistribution to servers or
  lists, or reuse of any copyrighted component of this work in other works.}
\newcommand\copyrightnotice{%
\begin{tikzpicture}[remember picture,overlay]
\node[anchor=south,yshift=10pt] at (current page.south) {\fbox{\parbox{\dimexpr\textwidth-\fboxsep-\fboxrule\relax}{\copyrighttext}}};
\end{tikzpicture}%
}

\begin{document}

\maketitle
\copyrightnotice
\thispagestyle{empty}
\pagestyle{empty}

\begin{abstract}

Event-based camera has emerged as a promising paradigm for robot perception, offering advantages with high temporal resolution, high dynamic range, and robustness to motion blur. However, existing deep learning-based event processing methods often fail to fully leverage the sparse nature of event data, complicating their integration into resource-constrained edge applications. While neuromorphic computing provides an energy-efficient alternative, spiking neural networks struggle to match of performance of state-of-the-art models in complex event-based vision tasks, like object detection and optical flow. Moreover, achieving high activation sparsity in neural networks is still difficult and often demands careful manual tuning of sparsity-inducing loss terms. Here, we propose Context-aware Sparse Spatiotemporal Learning (CSSL), a novel framework that introduces context-aware thresholding to dynamically regulate neuron activations based on the input distribution, naturally reducing activation density without explicit sparsity constraints. Applied to event-based object detection and optical flow estimation, CSSL achieves comparable or superior performance to state-of-the-art methods while maintaining extremely high neuronal sparsity. Our experimental results highlight CSSL's crucial role in enabling efficient event-based vision for neuromorphic processing.

\end{abstract}
\section{Introduction}

Event camera has emerged as a promising paradigm for visual perception in robotics, offering advantages such as high temporal resolution, high dynamic range, and robustness to motion blur compared to conventional frame-based cameras~\cite{gallego2020event}. These attributes make event cameras well-suited for real-time applications in dynamic environments, such as autonomous driving~\cite{Gehrig24nature} and drone-based agile flight~\cite{bhattacharya2024monoculareventbasedvisionobstacle}. However, despite these advantages, many existing methods still treat event-based vision as a conventional visual processing task, failing to fully exploit the inherent sparsity of event data~\cite{7539039}, which can be a limiting factor for real-time robotics applications, especially on resource-constrained platforms.

Neuromorphic computing offers an alternative paradigm for processing event data, leveraging the inherent sparsity of event streams to achieve high energy efficiency~\cite{yik2025neurobench}. Therefore, to fully take advantage of neuromorphic computing, it's necessary for the neural network model to maintain neuron activation sparsity. While sparse computing on neuromorphic platforms is promising, training sparse networks can be challenging and often requires careful tuning of loss functions to achieve satisfactory performance~\cite{NEURIPS2018_82f2b308,xu2024eventbasedopticalflowneuromorphic}. To bridge this gap, recent approaches have introduced sparse convolutional recurrent learning, which maintains activation sparsity in recurrent neural networks while preserving the representational capacity~\cite{wang2025sparse}.

Another promising method to maintain activation sparsity while keeping robust feature learning is introducing a self-adaptive mechanism, that regulates neuron action dynamically based on input characteristics. The context-awareness mechanism in neural networks can dynamically modulate network parameters, activations, or connections based on contextual features extracted from the input~\cite{bengio2015conditional}, which contrasts with traditional networks where the processing remains fixed regardless of the input. These mechanisms allow the network to adapt its processing strategy to focus on the most relevant information in the input, improving performance, efficiency, or robustness. Since the relevant information within an event stream can vary significantly depending on the motion patterns, lighting changes, and object interactions, context awareness mechanism aligns well with event-based vision. 

\begin{figure}[t]
    \centering
    \includegraphics[width=1.0\linewidth]{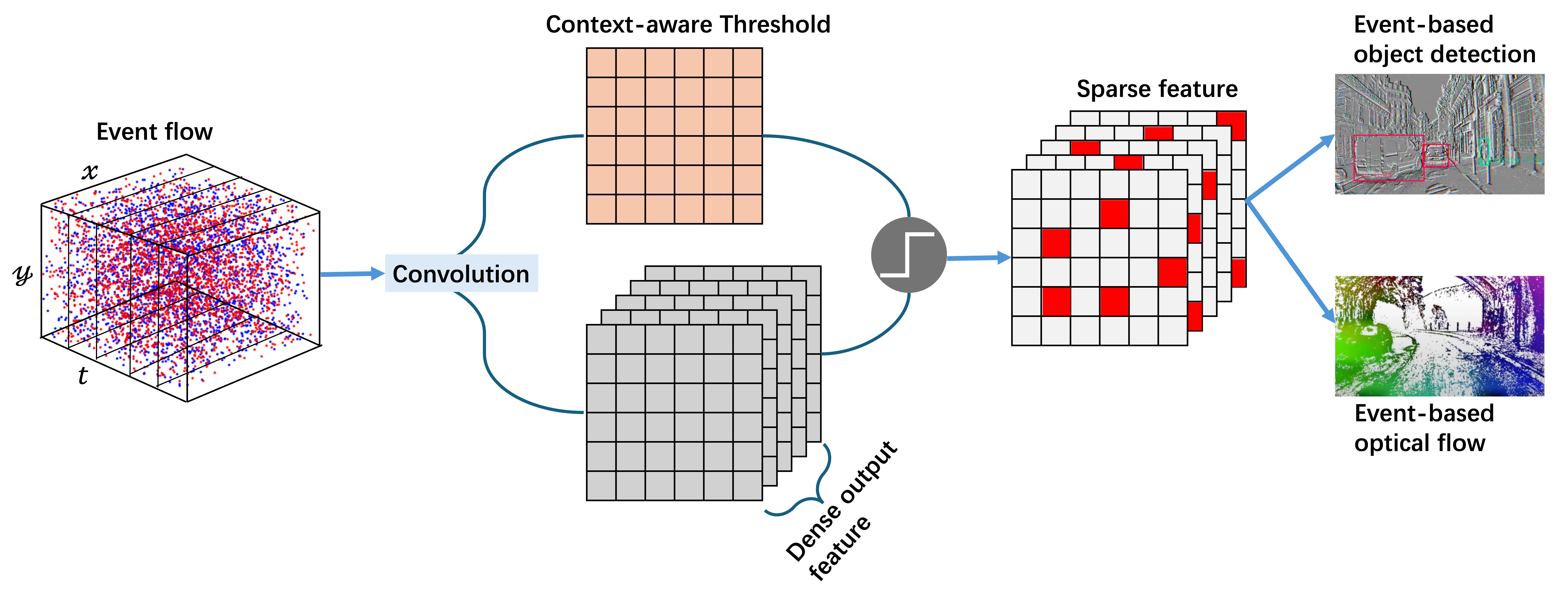}
    \caption{An overview of the proposed Context-aware Sparse Spatiotemporal Learning (CSSL) framework. CSSL introduces context-aware thresholding to dynamically regulate activations in convolutional modules, selectively filtering out redundant activations while preserving essential information. The framework is applied to event-based object detection and optical flow estimation.}
    \label{fig:Overview of CSSL for event-based vision}
\end{figure}

Inspired by recent advances in context-aware neural networks~\cite{pmlr-v202-liu23am}, we proposed Context-aware Sparse Spatiotemporal Learning (CSSL), a novel framework that dynamically modulates neuron activations based on contextual information to address these challenges. Unlike standard activation functions such as ReLU, which apply fixed thresholds, CSSL learns adaptive thresholds based on the input distribution, ensuring that only the most relevant neurons are activated. By incorporating context-aware thresholding into both convolutional and recurrent architectures, CSSL maximizes computational efficiency while improving task-specific performance. 

Our key contributions are as follows:

\begin{itemize}
    \item Context-aware sparse spatiotemporal learning framework: We introduced a novel context-aware learning approach that dynamically adjusts activation thresholds based on contextual information, enhancing spatial sparsity while preserving key event-driven features.
    \item Efficient spatiotemporal learning: We extended context-aware thresholding to both convolutional layers and convolutional recurrent units, enabling efficient processing of event-based data.
    \item Generalization to multiple tasks: We applied CSSL to event-based object detection and optical flow estimation, demonstrating state-of-the-art computation efficiency with minimal performance loss.
    \item Computational efficiency: Our experiment results show that CSSL significantly reduces computational overhead while maintaining accuracy, making it well-suited for real-time robotic applications.
\end{itemize}

\section{Methods}
\subsection{Overview of CSSL}

Event-based convolution is a specialized approach designed for sparse, asynchronous data processing in neuromorphic systems. Unlike standard convolutional operations, which process all pixels in a frame simultaneously, event-based convolution operates on individual spikes (events), significantly reducing computational redundancy and improving efficiency. Traditional frame-based convolutions store and process all pixels uniformly, requiring large memory overheads. In contrast, event-driven depth-first convolution prioritizes processing events as they arrive, integrating information incrementally and releasing memory as soon as it is no longer needed. This technique ensures efficient computation while keeping memory requirements low, making it particularly well-suited for event-based vision applications with dynamic and sparse input patterns.

Building upon event-based convolution~\cite{10.3389/fnins.2024.1335422}, we proposed Context-aware Sparse Spatiotemporal Learning (CSSL), a novel framework that introduces context-aware thresholding to dynamically modulate the activation of 2D convolution blocks. To enhance sparsity within convolution computing, we introduce this context-aware thresholding mechanism that adaptively filters activations based on input contextual features. Unlike traditional activation functions (e.g. ReLU) that apply a fixed threshold, our method dynamically determines the threshold value based on the input distribution. The threshold is derived by applying convolution to the input feature, ensuring that its spatial dimensions remain consistent with those of the output feature. The CSSL approach is highly flexible and can be applied to both standard convolutional layers and convolutional recurrent units, making it a generalizable solution for event-based visual processing. 

\subsection{Context-aware Threshold for 2D Convolution}

Compared to the normal 2D convolution layer generating dense hidden output feature, the context-aware thresholding mechanism generates key events to keep output sparse. Specifically, given an input feature map $x$, the threshold $v_{th}^{(t)}$ is computed as Equation~\ref{equation:Context-aware Threshold conv}:
\begin{equation}
    \label{equation:Context-aware Threshold conv}
    \begin{split}    
    v_{th}^{(t)} &= \sigma\left(W_vx^{(t)} + b_v\right)\\
    \tilde{y}^{(t)} &=W_xx^{(t)} + b_x\\
    s^{(t)} &= \text{H}\left(\tilde{y}^{(t)} - v_{th}^{(t)}\right)\\
    y^{(t)} &= s^{(t)}\odot \tilde{y}^{(t)}
    \end{split}
\end{equation}

where $W_v$ and $b_v$ denote convolution kernel and bias which computes threshold from input feature $x^{(t)}$. The terms $W_x$ and $b_x$ denote the convolution parameters that process the input feature into an intermediate representation $\tilde{y}^{(t)}$. The symbol $\odot$ denotes element-wise product and $\sigma(\cdot)$ denotes the sigmoid function ensuring that the computed threshold $v_{th}^{(t)}$ are projected into the normalized range of [0,1] and the sparse output feature is positive. The context-aware convolution unit dynamically regulates activations by applying a pixel-wise threshold to the processed feature map. As shown in Figure~\ref{fig:Context-aware threshold for convolution}, the event activation mask $s^{(t)}$ is generated using a Heaviside step function $H(\cdot)$ every time step which selectively retains informative features while filtering out less relevant activations. This mechanism enhances spatial sparsity by suppressing unnecessary computations while preserving meaningful event-based information. We used a surrogate gradient to estimate the backpropagated gradient of $H(\cdot)$ using the same methods presented in~\cite{subramoney2022efficient}. 

Given that event-based inputs are inherently spatially sparse, maintaining a dense output across the network is both computationally inefficient and unnecessary for effective feature learning. Instead of applying a uniform threshold across all activations, our proposed pixel-wise context-aware thresholding dynamically adjusts the activation threshold at each spatial location based on the local input context. This ensures that only informative event-driven activations are retained, while irrelevant or noisy activations are suppressed. By leveraging this fine-grained sparsity control, our method enhances computational efficiency and optimizes feature representation, making it particularly well-suited for event-based vision tasks where the relevant information is non-uniformly distributed across the input space.

\subsection{Context-aware Threshold for Residual block}

We integrate context-aware thresholding with residual block as shown in Figure~\ref{fig:Context-aware threshold for residual}. To accommodate the sparse input feature in the second convolutional layer, the first convolutional layer is replaced with a context-aware convolution. The output from the second convolution is then split channel-wise into two components: a threshold and a dense output feature. The dense output feature is accumulated with input $x^{(t)}$ with the residual connection. Unlike conventional residual blocks, which apply neuron-wise additions between the input and processed feature, our method introduces an additional thresholding step after accumulation. This ensures that the final output remains sparse, effectively reducing redundant activations.

\subsection{Context-aware Threshold for Event-based Convolutional Recurrent Learning}
\label{sec:context-aware Event-based Convolutional Recurrent Learning}

\begin{figure*}[htbp]
    \label{fig:Context-aware arch}
    \centering
    \subfigure[Context-aware threshold for convolution layer]{
        \label{fig:Context-aware threshold for convolution}
        \includegraphics[width=0.25\textwidth]{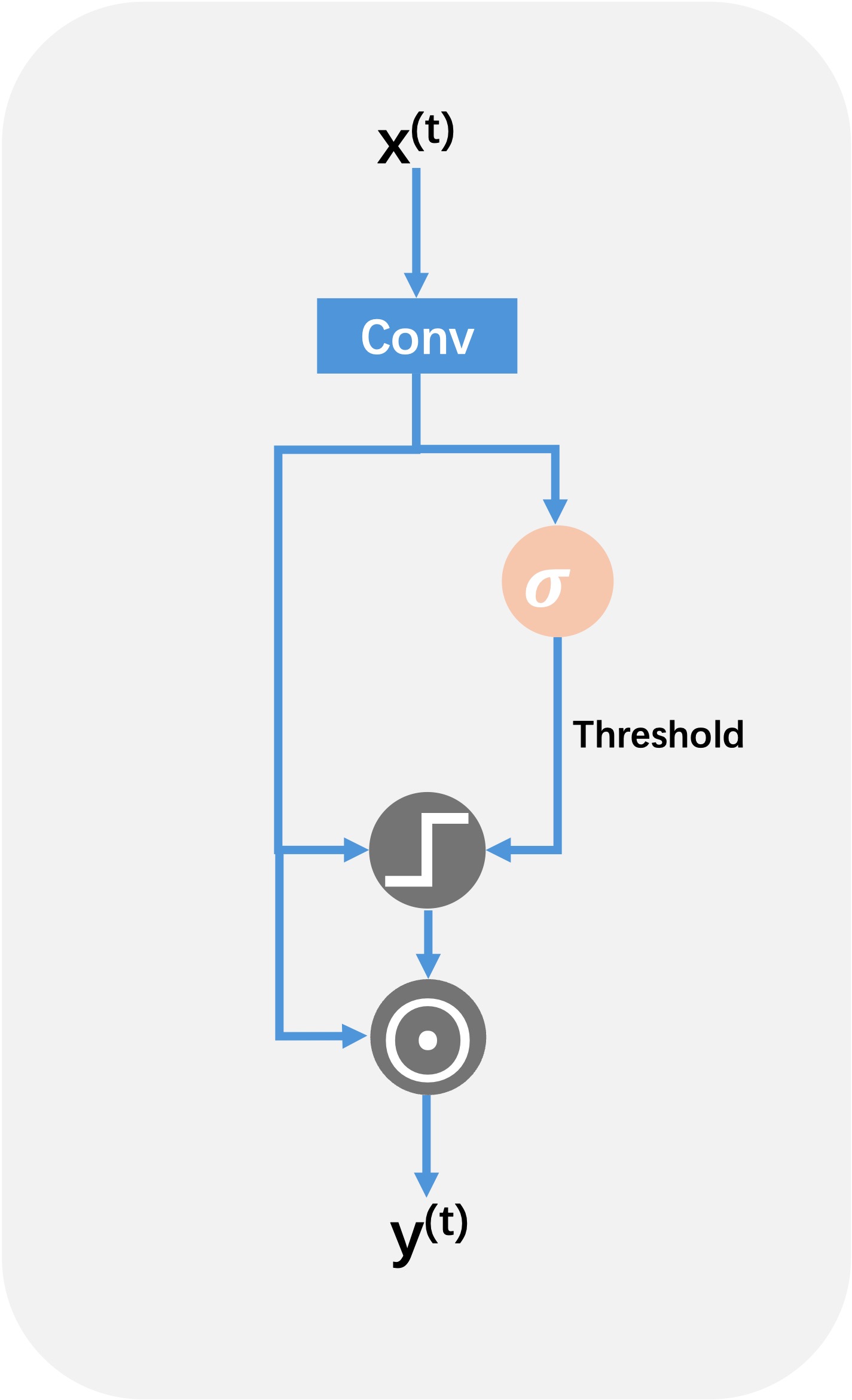}
    }
    \hfill
    \subfigure[Context-aware threshold for residual block]{
        \label{fig:Context-aware threshold for residual}
        \includegraphics[width=0.25\textwidth]{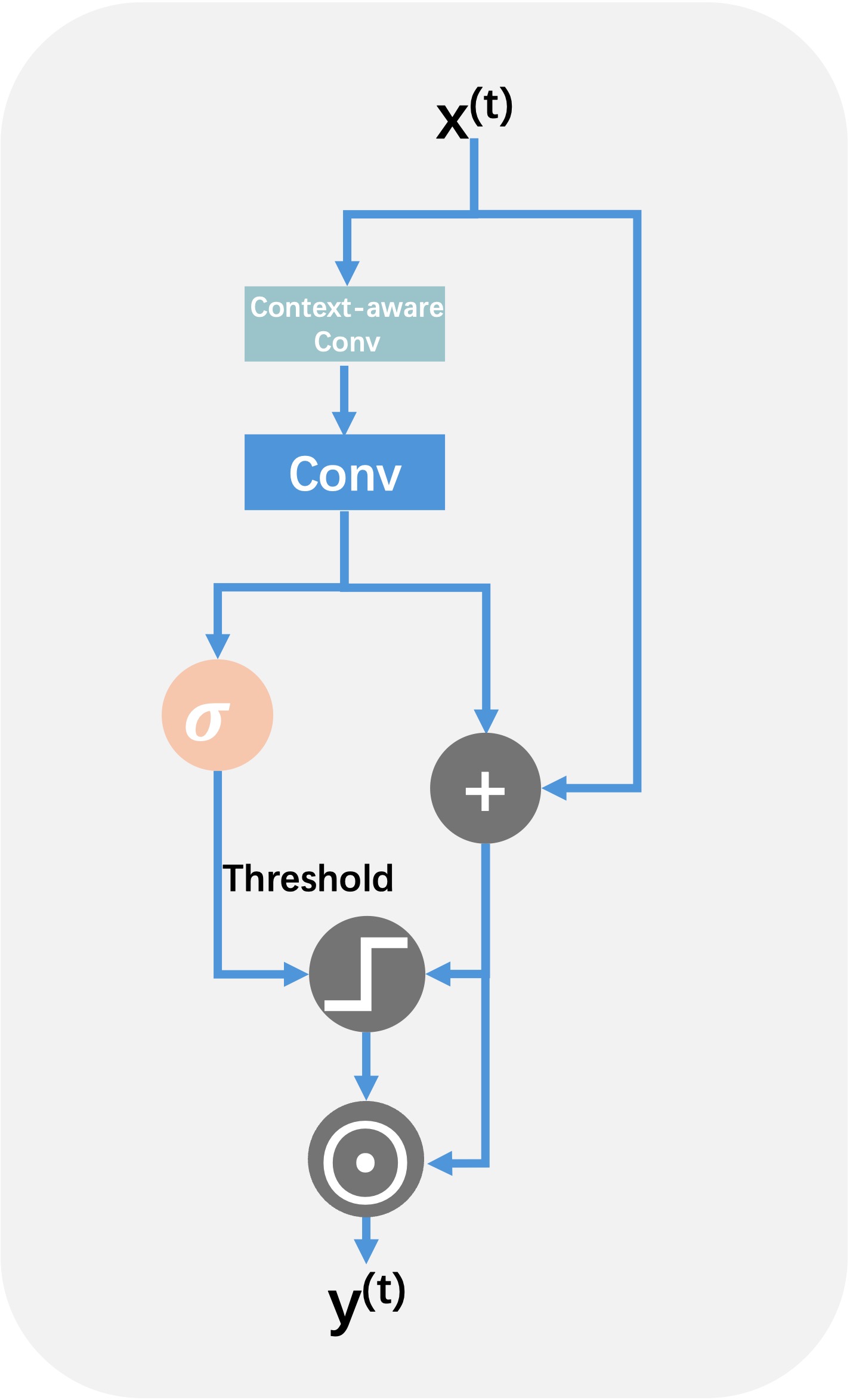}
    }
    \hfill
    \subfigure[Context-aware threshold for Conv-recurrent block]{
        \label{fig:Context-aware threshold for recurrent}
        \includegraphics[width=0.26\textwidth]{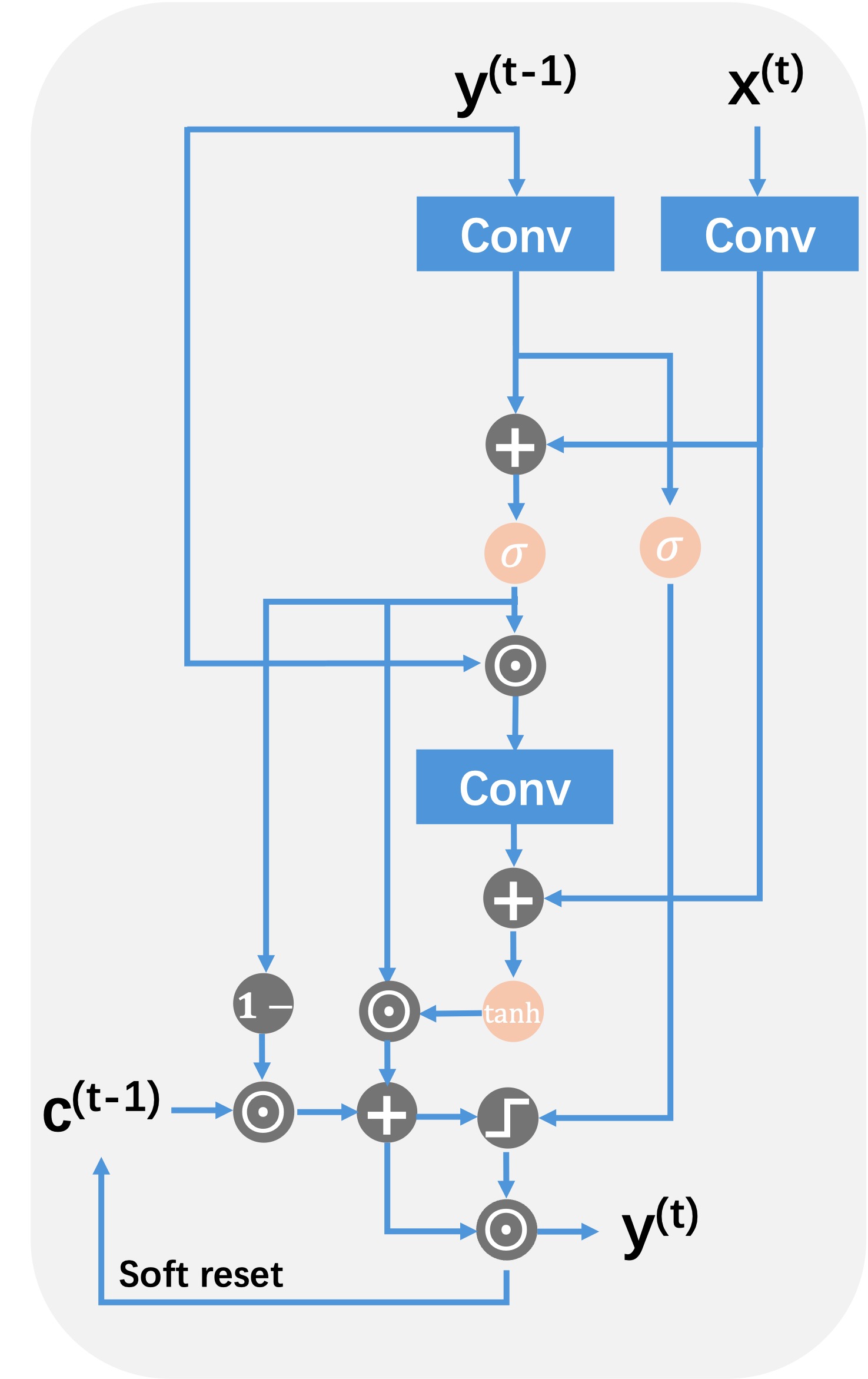}
    }
    \caption{Illustration of the context-aware thresholding process. The input feature is processed through a context-aware convolution that generates an adaptive threshold. This threshold selectively filters activations, allowing only informative signals to propagate while suppressing irrelevant ones. Compared to traditional fixed-threshold activation functions (e.g., ReLU), CSSL dynamically adjusts the threshold based on input distribution, enhancing spatial sparsity in convolutional layers and spatiotemporal sparsity in recurrent architectures.}
\end{figure*}

To explore the spatiotemporal sparsity in neural network learning, we extended and generalized our context-aware thresholding to the sparse convolutional recurrent unit~\cite{wang2025sparse}. As illustrated in Figure~\ref{fig:Context-aware threshold for recurrent}, we primarily employed the Minimal Gated Unit (MGU)~\cite{zhou2016minimal} for our experiments within the context-aware convolution recurrent unit, following the event recurrent processing method in~\cite{subramoney2022efficient}. The context-aware thresholding can also be applied to other sparse convolutional recurrent units using the same approach.

As shown in Equation~\ref{equation:Context-aware Threshold recurrent}, the threshold is obtained by applying convolution to the sparse hidden state from the previous time step $y^{(t-1)}$. This design choice is motivated by the fact that in event-based vision, objects can remain stationary, causing the event camera to cease generating new events. In such scenarios, a threshold derived from the memory cell ensures that the network selectively retains relevant activations, allowing it to produce accurate outputs even in the absence of new input events.

Since the output features resulting from the convolution of the input $x^{(t)}$ do not participate in subsequent convolutions, the dense tensor has a negligible impact on the total number of synaptic operations (SOp). Standard convolution can be used without thresholding mechanism. We added an auxiliary hidden state $c^{(t)}$ to the unit, and $y^{(t)}$ is generated in event-based form as sparse output of the unit. 

\begin{subequations}
\label{equation:Context-aware Threshold recurrent}
\begin{align}
    f^{(t)} &= \sigma\left(W_{xf}x^{(t)} + W_{yf}y^{(t-1)} + b_{f}\right) \\
    v_{th}^{(t)} &= \sigma\left(W_{v}y^{(t-1)} + b_{v}\right) \label{equ:vth}\\ 
    \tilde{h}^{(t)} &= \text{tanh}\left(W_{hi}\left(f^{(t)}\odot y^{(t-1)}\right) + W_{xh}x^{(t)} + b_{h}\right) \\
    c^{(t)} &= \left(1-f^{(t)}\right) \odot c^{(t-1)} + f^{(t)} \odot \tilde{h}^{(t)} \label{equ:ct}\\ 
    s^{(t)} &= \text{H}\left(c^{(t)} - v_{th}^{(t)}\right) \label{equ:event} \\ 
    y^{(t)} &= c^{(t)}\odot s^{(t)} \label{equ:output} 
\end{align}
\end{subequations}

Inspired by spiking neurons~\cite{guo2022reducing} to help to forget, we add a soft-reset mechanism to $c^{(t)}$ after the conv-rec unit generating $y^{(t)}$ as shown in Equation~\ref{equation:Context-aware Threshold}:
\begin{equation}
    \label{equation:Context-aware Threshold}
    c^{(t)} = c^{(t)} - v_{th}^{(t)} \odot s^{(t)}
\end{equation}

By dynamically adjusting the hidden state based on the learned context-aware threshold, network maintains sparse yet informative representations over time. 

\subsection{Apply to robotics application}

To evaluate the generalizability of the proposed CSSL framework, we applied it to two fundamental event-based vision tasks: object detection and optical flow estimation. By incorporating CSSL into different neural architectures, we evaluated its ability to enhance computational efficiency and task performance across diverse robotic perception applications. Specifically, we replaced standard convolution layers with context-aware convolutions, introduced context-aware thresholding in residual and recurrent blocks.

Our experiments are designed to achieve two primary objectives. First, we demonstrated that simply replacing standard convolutional units with our context-aware modules leads to significant performance improvements across different tasks. Second, we showed that our context-aware approach enables the network to naturally achieve high sparsity and robust performance, even without the need for specialized sparsity regularization techniques.

\section{Experiments and Results}
\subsection{Experimental Setup}
\label{sec:Experimental Setup}

\subsubsection{Event-based Object Detection}

For the event-based object detection task, our method was evaluated on the 1 Mpx~\cite{perot2020learning} and Gen1~\cite{de2020large} datasets, adhering to the standard evaluation approach outlined in~\cite{perot2020learning}. Performance is reported by widely used COCO mAP metric~\cite{lin2014microsoft}. The CSSL framework incorporated MGU within Conv-Rec modules, except in Section~\ref{sec:Different recurrent unit}, where alternative recurrent unit were explored. The best-performing model on the validation set was subsequently applied to the test set to obtain the final mAP score.

We validated the proposed method using the Sparse Event-based Efficient Detector (SEED) architecture~\cite{wang2025sparse}, which includes the backbone and the SSD detection head~\cite{liu2016ssd}. The architecture of the backbone is shown in Table~\ref{tab:Backbone Architecture of SEED-256}. The CSSL-SEED network backbone comprises three key components: the context-aware convolution module, the context-aware residual block, the context-aware conv-rec block.

\begin{table}[htbp]
\caption{Backbone Architecture of SEED-256}
    \label{tab:Backbone Architecture of SEED-256}
  \begin{center}
    \begin{tabular}{ccc}
    \hline
    Layer & Channels out & Dimension out(1Mpx)  \\\hline
    CSSL Conv & 32 & [320, 180]  \\
    CSSL Residual Connection & 64 & [160, 90] \\
    CSSL Residual Connection & 64 & [160, 90]  \\
    CSSL Residual Connection & 128 & [160, 90]  \\
    CSSL ConvMGU & 256 & [80, 45]  \\
    CSSL ConvMGU & 256 & [40, 23]  \\
    CSSL ConvMGU & 256 & [20, 12]  \\\hline
    
    \end{tabular}
    \end{center}

\end{table}

Using ADAM optimizer, all models are trained with full precision with the OneCycle scheduler to adjust the learning rate dynamically. For 1Mpx dataset, models are trained for 50 epochs with the maximum learning rate of 3e-4 and for Gen1 datasets, models are trained for 60 epochs with the maximum learning rate set at 2.5e-4. Using a NVIDIA A100 GPU, we trained our models with a batch size of 8, sequence length of 15 on 1Mpx dataset. The training takes approximately 4 days on a single A100 GPU. We trained our models with a batch size of 5, sequence length of 40 on the Gen1 dataset. The training takes approximately 3 days on a single A100 GPU.

\subsubsection{Event-based Optical Flow}

For the event-based optical flow estimation task, we adopted the same neural network architecture as EV-FlowNet in~\cite{xu2024eventbasedopticalflowneuromorphic}. We integrated our context-aware modules by replacing standard convolution layers with context-aware convolution, conventional residual blocks with context-aware residual blocks, and ConvGRU with context-aware MGU blocks. Performance was assessed using Average Endpoint Error (AEE) and outlier percentage metrics to evaluate both accuracy and robustness.

We followed the experimental setup outlined in~\cite{xu2024eventbasedopticalflowneuromorphic}. The network architecture, training strategies, and evaluation protocols remain consistent with~\cite{xu2024eventbasedopticalflowneuromorphic}. The AdamW optimizer was employed, with a OneCycle learning rate scheduler dynamically adjusting the learning rate during training, with a maximum learning rate of 2e-4. The model was trained for 100 epochs on the UZH-FPV dataset~\cite{delmerico2019we}. Inference and evaluation were conducted on the MVSEC dataset~\cite{zhu2018multivehicle}. The training takes approximately 10 hours on a single RTX4090 GPU.

\subsection{Experiments results}

We benchmarked our model against state-of-the-art methods for event-based vision in Table~\ref{tab:main-result} for object detection and Table~\ref{tab:Result of optical flow} for optical flow estimation.

In the event-based object detection task, CSSL significantly reduces synaptic operations while preserving high detection accuracy. As shown in Table~\ref{tab:main-result}, the CSSL-based models outperform conventional architectures in terms of computational efficiency, achieving comparable or superior mean Average Precision (mAP) scores with significantly lower GSOp. Notably, the CSSL-MGU model achieves an mAP of 46.4 on the 1Mpx dataset  while requiring only 2.8 GSOp, which is only 32.2\% of RVT-S~\cite{gehrig2023recurrent}. Compared to SNN-based solutions, our model outperforms state-of-the-art methods in mAP while requiring only 7.4\% of the synaptic operations. This result indicates that CSSL effectively balances accuracy and efficiency by dynamically filtering activations, thereby reducing redundant computations in dense event-based input streams.

\begin{table*}[htbp]
\caption{Comparing CSSL-SEED with state-of-the-art object detection approaches for event-based vision}
  \label{tab:main-result}
  \vskip 0.15in
  \begin{center}
\begin{tabular}{ccccccc}
\hline
 \multicolumn{2}{c|}{} & \multicolumn{2}{c|}{1 Mpx} & \multicolumn{2}{c|}{Gen1} &  \\ 
\hline
Method & \multicolumn{1}{c|}{Network} & mAP  & \multicolumn{1}{c|}{GSOp} & mAP  & \multicolumn{1}{c|}{GSOp} & Param(M)\\
\hline
ASTMNet~\cite{li2022asynchronous} & \multicolumn{1}{c|}{TACN+ConvRec+SSD} & 48.3  & \multicolumn{1}{c|}{-} & 46.7  & \multicolumn{1}{c|}{-} & $>$39.6\\
SNN~\cite{cordone2022object} & \multicolumn{1}{c|}{Spiking DenseNet+SSD} & -  & \multicolumn{1}{c|}{-} & 18.9  & \multicolumn{1}{c|}{2.33} & 8.2\\
SpikeYOLO~\cite{10.1007/978-3-031-73411-3_15} & \multicolumn{1}{c|}{Spiking+YOLOv8} & - & \multicolumn{1}{c|}{-} & 40.4 & \multicolumn{1}{c|}{14.3} & 23.1\\
RED~\cite{perot2020learning} & \multicolumn{1}{c|}{SENet+ConvLSTM+SSD} & 43.0  & \multicolumn{1}{c|}{26.1} & 40.0 & \multicolumn{1}{c|}{8.26} & 24.1 \\
RVT-B~\cite{gehrig2023recurrent} & \multicolumn{1}{c|}{MaxViT+LSTM+YOLOX} & 47.4 & \multicolumn{1}{c|}{15.6} & 47.2 & \multicolumn{1}{c|}{5.05} & 18.5\\
RVT-S~\cite{gehrig2023recurrent} & \multicolumn{1}{c|}{MaxViT+LSTM+YOLOX} & 44.1 & \multicolumn{1}{c|}{8.69} & 46.5 & \multicolumn{1}{c|}{2.78} & 9.9\\
RVT-T~\cite{gehrig2023recurrent} & \multicolumn{1}{c|}{MaxViT+LSTM+YOLOX} & 41.5 & \multicolumn{1}{c|}{3.87} & 44.1 & \multicolumn{1}{c|}{1.29} & 4.4\\
SEED-256~\cite{wang2025sparse} & \multicolumn{1}{c|}{ECNN+EConvGRU+SSD} & 44.9  & \multicolumn{1}{c|}{3.83} & 45.3 & \multicolumn{1}{c|}{1.32} & 13.9 \\
SEED-128~\cite{wang2025sparse} & \multicolumn{1}{c|}{ECNN+EConvGRU+SSD} & 44.1  & \multicolumn{1}{c|}{2.75} & 44.5 & \multicolumn{1}{c|}{0.99} & 4.8 \\
\hline
\textbf{CSSL-SEED-256} & \multicolumn{1}{c|}{CSSL-Conv+CSSL-ConvMGU+SSD} & 46.4  & \multicolumn{1}{c|}{2.80} & 46.3 & \multicolumn{1}{c|}{1.06} & 10.7 \\
\textbf{CSSL-SEED-256} & \multicolumn{1}{c|}{CSSL-Conv+CSSL-ConvGRU+SSD} &  46.2 & \multicolumn{1}{c|}{3.42} & 46.4 & \multicolumn{1}{c|}{1.22} &  13.9\\
\textbf{CSSL-SEED-256} & \multicolumn{1}{c|}{CSSL-Conv+CSSL-ConvMinimalRNN+SSD} & 44.8  & \multicolumn{1}{c|}{2.71} & 45.5 & \multicolumn{1}{c|}{0.99} &  7.9\\
\hline
\end{tabular}
  \end{center}

\end{table*}

Similarly, in the event-based optical flow estimation task, our CSSL-integrated models demonstrate superior computational efficiency compared to conventional recurrent-based architectures. As summarized in Table~\ref{tab:Result of optical flow}, the fully CSSL-integrated CSSL-EV-FlowNet achieves an AEE of 2.38, outperforming RNN-EV-FlowNet~\cite{xu2024eventbasedopticalflowneuromorphic} with specialized FATReLU sparsification, and only 62.8\% neuron density compared to it. 

Our experimental results demonstrate the effectiveness of CSSL in improving both computational efficiency and task performance across event-based object detection and optical flow estimation tasks. The integration of context-aware thresholding in convolutional and recurrent modules enables our models to selectively retain informative activations, reducing computational redundancy while maintaining accuracy. The sparsity-driven approach of CSSL allows for a more efficient allocation of computational resources, making it highly suitable for real-time robotic applications where power and memory efficiency are critical.

\begin{table*}[htbp]
\caption{Comparing CSSL-EV-FlowNet with state-of-the-art optical flow estimation approaches for event-based vision}
\label{tab:Result of optical flow}
\begin{center}
    \begin{tabular}{cccccccccccc}
    \hline
    \multirow{2}{*}{Network} & \multicolumn{2}{c}{outdoor\_day1} & \multicolumn{2}{c}{indoor\_flying\_1} & \multicolumn{2}{c}{indoor\_flying2} & \multicolumn{2}{c}{indoor\_flying3} & \multicolumn{3}{c}{Average} \\ 
    \cmidrule(lr){2-3} \cmidrule(lr){4-5} \cmidrule(lr){6-7} \cmidrule(lr){8-9} \cmidrule(lr){10-12}
                             & AEE             & $\%_{outlier}$             & AEE               & $\%_{outlier}$               & AEE              & $\%_{outlier}$              & AEE              & $\%_{outlier}$              & AEE     & $\%_{outlier}$     & Dens.(\%)   \\ \hline
    EV-FlowNet(GRU)~\cite{hagenaars2021self}          & 1.69            & 12.50           & 2.16              & 21.51             & 3.90             & 40.72            & 3.00             & 29.60            & 2.94    & 29.35   &     -    \\
    RNN-EV-FlowNet~\cite{xu2024eventbasedopticalflowneuromorphic}           & 1.69            & 12.96           & 2.02              & 18.74             & 3.84             & 38.17            & 2.97             & 27.91            & 2.88    & 27.32   & 16.90   \\
    \textbf{CSSL-EV-FlowNet}        & 1.55            & 11.64          & 1.84             & 14.84             & 3.43             & 33.77           & 2.68            & 25.01            & 2.38    & 21.31  & 10.61 \\ 
    \hline
    \end{tabular}
\end{center}

\end{table*}

\subsection{Generalizing to variations of recurrent unit}
\label{sec:Different recurrent unit}

We extended the CSSL method to various recurrent architectures to validate its generalizability, specifically applying it to MinimalRNN~\cite{chen2017minimalrnn} and GRU~\cite{cho2014learning}. These context-aware recurrent units, similar to the approach described in Section~\ref{sec:context-aware Event-based Convolutional Recurrent Learning}, dynamically generate pixel-wise thresholds from the output feature of the convolution applied to the previous hidden state $y^{(t-1)}$ at the current time step, as formulated in Equation~\ref{equ:vth}. Additionally, we introduced an auxiliary hidden state $c^{(t)}$ in Equation~\ref{equ:ct} and applied the thresholding mechanism to regulate its updates, ensuring the generation of events, as described in Equation~\ref{equ:event}. However, we did not extend our method to LSTM due to its additional memory cell, which introduces a separate gating mechanism. This additional memory cell complicates the direct integration of our context-aware thresholding approach, making it less compatible with the sparsity-driven processing strategy employed in CSSL.

We evaluated our method on the event-based object detection task on the 1Mpx and Gen1 datasets, with the results presented in Table~\ref{tab:main-result}. Notably, GRU achieves comparable performance to MGU while requiring slightly higher synaptic operations and parameter count. In contrast, MinimalRNN exhibits lower performance compared to both MGU and GRU but benefits from a reduced model size, highlighting a trade-off between efficiency and accuracy.

\subsection{Effectiveness of context-aware sparse learning}

To assess the effectiveness of our proposed context-aware thresholding method, we compared it with a baseline model that employs standard ReLU activation in all straight-forward convolutional layers and incorporates a sparsity loss function, as shown in Equation~\ref{equation:sparse_loss}, where $\beta_{sparse}$ is the weighting coefficient of the sparsity loss, $n$ represents the index of a training sample, $N$ denotes the mini-batch size, $t$ corresponds to the step index within a training sample, $T$ represents the total number of steps per sample, $l$ indicates the index of the layer, $L$ is the overall number of layers, and $\mathbf{x}$ refers to the activation maps.

\begin{equation}
\label{equation:sparse_loss}
    L_{sparse}=\beta_{sparse}\frac{1}{NT}\sum_{n=1}^{N}\sum_{t=1}^{T}\sum_{l=1}^{L} ||\mathbf{x}^{(n)(t)(l)}||_{1}
\end{equation}

We employed a two-stage training strategy to evaluate the impact of sparsity constraints on model performance and efficiency. The first stage follows the same setup as described in Section~\ref{sec:Experimental Setup}, with training conducted without the sparsity loss constraint. In the second stage, we introduced the sparsity loss and vary its values to assess its effect on performance. Throughout all experiments, we used MGU as the recurrent unit for all convolution recurrent modules. The results are summarized in Table~\ref{tab:sparse result}.

\begin{table}[htbp]
    \centering
    \caption{Results with sparsity loss}
    \label{tab:sparse result}
    \begin{tabular}{ccccc}
        \hline
          & \multicolumn{4}{|c}{1Mpx}                               \\ \hline
          & \multicolumn{2}{|c}{Conv + ReLU + L1}    & \multicolumn{2}{c}{CSSL + L1} \\ 
        \cmidrule(lr){2-3} \cmidrule(lr){4-5} 
       \multicolumn{1}{c|}{$\beta_{sparse}$}    & mAP & GSOp                  & mAP        & GSOp        \\ \hline
\multicolumn{1}{c|}{1}     &  37.4   & \multicolumn{1}{c|}{1.85} &   40.0         &   1.54          \\
\multicolumn{1}{c|}{0.1}   &  44.8   & \multicolumn{1}{c|}{3.38} &   46.1         &   2.17          \\
\multicolumn{1}{c|}{0.04}  &  44.9   & \multicolumn{1}{c|}{5.18} &   46.3         &   2.63          \\
\multicolumn{1}{c|}{0.01}  &  45.1   & \multicolumn{1}{c|}{5.66} &   46.2         &   2.75          \\
\multicolumn{1}{c|}{0.001} &  44.3   & \multicolumn{1}{c|}{5.84} &   46.0         &   2.82       \\\hline
W/O $\beta_{sparse}$ & 45.8 & \multicolumn{1}{c|}{5.95} & 46.4 & 2.8 \\\hline
    \end{tabular}

\end{table}

\begin{table*}[htbp]
    \caption{Result comparison of average neural activation density of each SEED layer}
    \label{tab:layer density result}
    \begin{center}
    \begin{tabular}{l|ccc}
\hline
            & \multicolumn{3}{c}{Average Neural Activation Density (with mAP)}                                                                                                \\ \cline{2-4} 
            & \multicolumn{1}{c|}{CSSL without sparsity loss (46.4)} & \multicolumn{1}{c|}{SEED without sparsity loss (45.0)} & SEED + L1 (0.1) sparsity loss (44.2) \\ \hline
Conv1       & \multicolumn{1}{c|}{0.47}                       & \multicolumn{1}{c|}{0.66}                           &         0.52                      \\ \hline
Res1\_Conv1 & \multicolumn{1}{c|}{0.17}                       & \multicolumn{1}{c|}{0.61}                           &         0.53                     \\ \hline
Res1\_Conv2 & \multicolumn{1}{c|}{0.20}                       & \multicolumn{1}{c|}{0.72}                           &         0.45                      \\ \hline
Res2\_Conv1 & \multicolumn{1}{c|}{0.10}                       & \multicolumn{1}{c|}{0.56}                           &         0.50                      \\ \hline
Res2\_Conv2 & \multicolumn{1}{c|}{0.46}                       & \multicolumn{1}{c|}{0.84}                           &         0.66                      \\ \hline
Res3\_Conv1 & \multicolumn{1}{c|}{0.07}                       & \multicolumn{1}{c|}{0.43}                           &         0.36                      \\ \hline
Res3\_Conv2 & \multicolumn{1}{c|}{0.12}                       & \multicolumn{1}{c|}{0.50}                           &         0.38                      \\ \hline
Recurrent1  & \multicolumn{1}{c|}{0.08}                       & \multicolumn{1}{c|}{0.05}                           &         0.05                      \\ \hline
Recurrent2  & \multicolumn{1}{c|}{0.07}                       & \multicolumn{1}{c|}{0.05}                           &         0.04                      \\ \hline
Recurrent3  & \multicolumn{1}{c|}{0.06}                       & \multicolumn{1}{c|}{0.04}                           &         0.04                      \\ \hline
    \end{tabular}
    \end{center}
\end{table*}

First, we evaluated a baseline model that replaces our context-aware thresholding with a straightforward convolution that incorporates the ReLU activation function. As observed in Table~\ref{tab:sparse result}, this modification leads to a decrease in mAP performance in the 1Mpx dataset by 0.6, with synaptic operation increasing by 112.5\% compared to our proposed CSSL approach.  These findings indicate that our context-aware thresholding mechanism not only improves computational efficiency but also enhances the model’s ability to extract meaningful features. Additionally, we examined the effect of different $\beta_{sparse}$ values on training outcomes. The results highlight that the choice of $\beta_{sparse}$ significantly influences the detection performance. This underscores the advantage of our CSSL method, which achieves optimal performance in a single-stage training process without fine-tuning sensitive hyperparameters.

Furthermore, we investigated whether adding a sparsity loss directly to a pre-trained CSSL model could further reduce the number of synaptic operations. As shown in Table~\ref{tab:sparse result}, when setting $\beta_{sparse}$ to 0.04, the model maintains nearly the same performance while achieving a 6\% reduction in synaptic operations. However, when increasing $\beta_{sparse}$ further in an attempt to further decrease synaptic operations, we observed a notable performance degradation. These findings suggest that our proposed CSSL approach inherently achieves a near-optimal trade-off between performance and computational efficiency in the first-stage training itself, without requiring additional sparsity constraints. This demonstrates the effectiveness of our context-aware thresholding mechanism in maximizing performance while minimizing computation cost.

\section{Comparing per-layer activation density}

To further assess how context‑aware thresholding enhances activation sparsity, we measured the mean activation density for every layer of the SEED-256 architecture with and without using the proposed CSSL framework (Table~\ref{tab:layer density result}). The activation densities were averaged over the test split of the 1Mpx object‑detection dataset. Relative to the baseline SEED model trained with a sparsity loss, our proposed CSSL approach lowers the activation density in convolutional layers, without requiring an additional, compute‑intensive fine‑tuning phase.

\section{Discussion and Conclusion}
This paper presents CSSL, a computationally efficient framework for event-based vision. By applying context-aware thresholding mechanism, CSSL effectively reduces neuron activation density in convolutional modules, improving computation efficiency and task performance. Unlike traditional sparsity-driven networks that rely on explicit sparsity regularization, CSSL naturally achieves high activation sparsity by dynamically adapting thresholds based on the input distribution. This eliminates the need for manually tuned sensitive sparsity hyperparameters, making training more stable and efficient. Our approach directly addresses the challenge of high-cost spatiotemporal processing, setting a new benchmark for efficient event-based object detection and optical flow estimation for neuromorphic methods.

As neuromorphic processors increasingly support deep learning models~\cite{pei2019towards, liu2018memory, schuman2022opportunities}, CSSL is positioned to substitute spiking neural networks on neuromorphic processors with superior performance while maintaining low energy consumption.

Beyond object detection and optical flow, the principles of context-aware thresholding can be extended to various event-based tasks, including motion prediction, agile flying, and autonomous navigation. Integrating CSSL with recent advancements in event-based preprocessing~\cite{zubic2023chaos, peng2023better} could further enhance its adaptability across diverse applications. Additionally, the potential for ultra-fast inference with minimal neuron activation makes CSSL an ideal option for real-time edge deployment in high-speed robotics and automotive systems, where efficiency and responsiveness are important.

\bibliographystyle{IEEEtran}
\bibliography{myref}
\end{document}